# To What Extent Do Large Language Models Understand Bangla Idioms?

**Mousumi Akter[1,2], Md. Faiyaz Abdullah Sayeedi[3], Nurul Labib Sayeedi[4], Swakkhar Shatabda[3]**
[1]Technical University Dortmund [2]Research Center Trustworthy Data Science and Security
[3]BRAC University [4]United International University
Correspondence: mousumi.akter@tu-dortmund.de

## Abstract

Idiomatic expressions are an integral part of natural language, reflecting cultural nuances and posing unique challenges for computational models, particularly in low-resource languages. In this paper, we present the first large-scale benchmark dataset of Bangla idioms, complemented by a synthetic multiple-choice question (MCQ) dataset for idiom meaning identification. We conduct a comprehensive evaluation of recent large language models (LLMs) across three idiom-related tasks: paraphrasing, idiom span detection, and meaning identification, leveraging zero-shot and few-shot prompting strategies. Our results reveal substantial variability in model performance, with no single LLM consistently outperforming others across all tasks. Notably, *Phi-4-mini-instruct* excels in paraphrasing, *Kimi-K2-32b-instruct* in span detection, and *Gemini-2.5-flash* in meaning identification. We believe that our datasets and analyses will provide valuable resources to guide future research in improving LLM comprehension of idiomatic expressions, particularly in Bangla and other low-resource languages.[1]

## 1 Introduction

Idioms are multiword expressions whose meanings often diverge from their literal interpretation. Rooted in cultural and historical contexts, idioms are unique to specific linguistic communities. Bangla, the seventh most widely spoken language globally, is particularly rich in idiomatic expressions (an example is illustrated in Figure 1). However, the scarcity of annotated resources and the lack of a large-scale benchmark dataset for Bangla idioms significantly hinder progress in developing and evaluating robust NLP models (Joshi et al., 2020a; Chakraborty et al., 2021).

In this work, we present the first large-scale benchmark dataset of Bangla idioms, comprising

[1]The dataset will be made available upon acceptance.

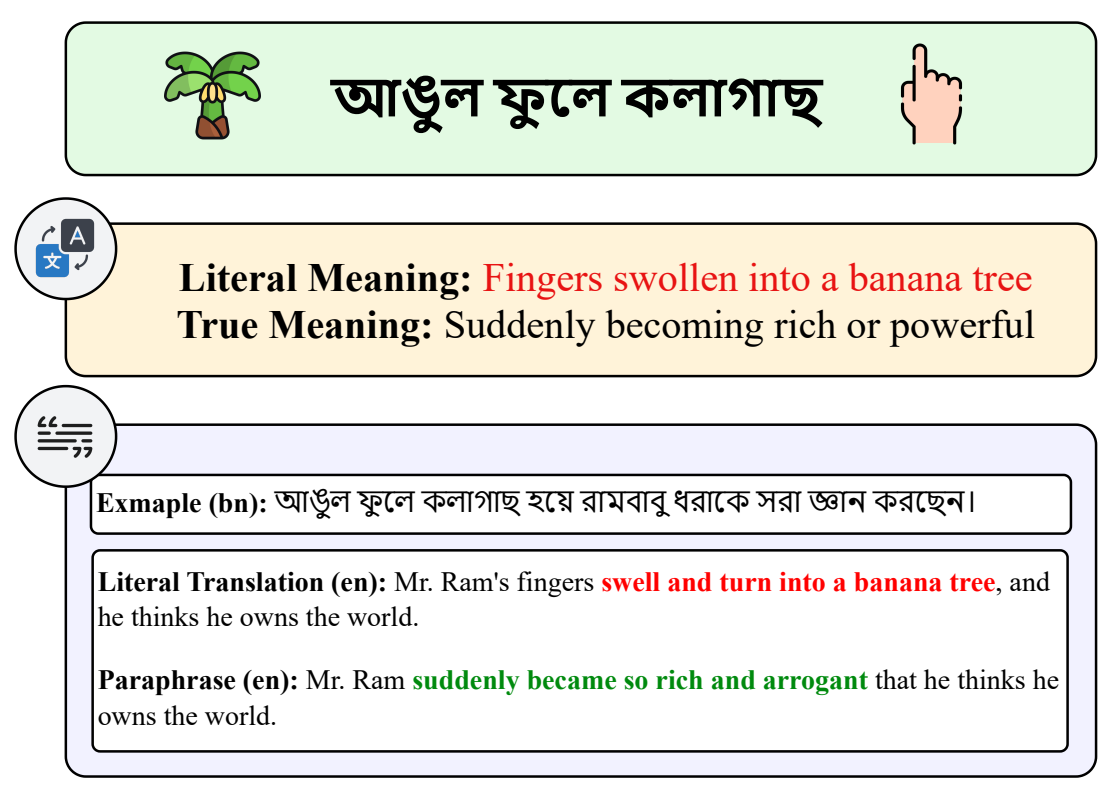

Figure 1: Example of the Bangla idiom "আঙুল ফুলে কলাগাছ" (*Suddenly becoming rich or powerful*), showing its literal translation vs. its true meaning.

**10,822** entries with their corresponding meanings, of which **4,772** additionally include usage examples. We further construct a synthetic multiple-choice question (MCQ) dataset for idiom meaning detection, consisting of **10,913** samples with single correct answers and **38,688** samples with multiple correct answers.

We additionally investigate the performance of recent large language models (LLMs) on three idiom-related tasks: paraphrasing, idiom span detection, and MCQ-based meaning detection. Our experiments cover several LLMs, including Kimi-K2-32B-Instruct (Bai et al., 2025), LLaMA-4-Scout-17B-Instruct (Meta, 2025), DeepSeek-R1-Distill-LLaMA-70B (Guo et al., 2025), GPT-OSS-20B, GPT-OSS-120B (OpenAI, 2025), Phi-4-Mini-Instruct (Abdin et al., 2024), Qwen3-32B (Team, 2024), and Gemini-2.5-Flash (Comanici et al., 2025). The results show that *Phi-4-mini-instruct* excels in paraphrasing, *Kimi-K2-32b-instruct* in idiom span detection, and *Gemini-2.5-flash* in meaning identification.

In Summary:

- We introduce the first large-scale bench-

mark dataset of Bangla idioms, consisting of 10,822 entries with meanings and 4,772 usage examples.

- We construct a large-scale synthetic MCQ dataset for idiom meaning detection with both single-answer (10,913 samples) and multiple-answer (38,688 samples) settings.
- We provide a comprehensive evaluation of recent large language models on three idiom-related tasks in Bangla, highlighting their strengths and limitations in low-resource idiomatic language understanding.

## 2 Related Works

**Idiom Understanding in NLP.** Idiomatic expressions, often represented as multiword expressions (MWEs), have been widely studied in NLP. Early work surveyed MWE processing (Constant et al., 2017) and released annotated corpora for Hindi and Marathi (Singh et al., 2016). Later resources include English (Zhou et al., 2021), multilingual Id10M (Tedeschi et al., 2022), and Chinese idiom datasets (Qiang et al., 2023). Idiomatic translation remains challenging (Dankers et al., 2022; Rezaeimanesh et al., 2025), and recent studies have analyzed idiomaticity in LLMs through sensitivity tests (Phelps et al., 2022; Klubička et al., 2023; Liu and Lareau, 2024; Yayavaram et al., 2024), hybrid identification (Alves et al., 2024), and commercial system evaluation (Phelps et al., 2024). Multilingual datasets exist for Indian languages (Agrawal et al., 2018) and other languages such as Catalan, Italian, and Russian (Sentsova et al., 2025). More recently, the PARSEME 2.0 Shared Task (Scholivet et al., 2026) has promoted multilingual idiom identification and paraphrasing across 14 languages.

**Bangla NLP and Our Contribution.** Bangla, spoken by over 300 million people, remains underrepresented in NLP (Joshi et al., 2020b), though recent datasets cover summarization (Hasan et al., 2021), QA (Bhattacharjee et al., 2022; Ekram et al., 2022), paraphrasing (Akil et al., 2022), back-transliteration (Fahim et al., 2024), and NER (Mahtab et al., 2025), etc. Recently, there has been small scale effort for figurative understanding of Bangla Idioms (Das et al., 2026; Sakhawat et al., 2026). To our knowledge, no prior work has systematically evaluated LLMs on Bangla idiom-related tasks (i.e., paraphrasing, span detection and MCQ meaning detection). We address this gap by introducing a large-scale Bangla idiom dataset and a synthetic MCQ corpus for meaning detection, and by benchmarking recent LLMs on paraphrasing, idiom span detection, and meaning identification, providing insights into their performance in a low-resource setting.

## 3 Bangla Idiom Dataset

We present a curated benchmark of Bangla idioms, manually collected from open-access and permissive sources under fair dealing provisions—including the National Curriculum and Text Book (NCTB) (বাংলা ভাষার ব্যকরন), Bangla Newspapers, Bangla Wikipedia[2], and online idiom lexicons[3]. More details are in Appendix A.2. The dataset underwent systematic cleaning to remove duplicates and apply minimal normalization (e.g., spacing corrections), while retaining authentic spelling variants without transliteration. Native Bangla-speaking authors manually verified that all examples reflect genuine idiomatic usage with semantically coherent meanings. To ensure legal compliance and transparency, a full source-by-source licensing mapping is detailed in Appendix A.1. We will distribute the benchmark dataset under a CC BY-SA 4.0 license upon acceptance.

| Field | Example |
|---|---|
| idiom | অমাবস্যার চাঁদ (*moon of the new moon night*) |
| meaning | অদর্শনীয় বস্তু, দুর্লভ বস্তু, খুব কম দেখা যায় (*something seldom seen; a rare thing; seen very rarely*) |
| sentence | আজকাল তোমাকে দেখাই যায় না, চাকরি পেয়ে কি অমাবস্যার চাঁদ হয়ে গেলে? (*We hardly see you these days, is getting a job make you moon of the new moon night?*) |

Table 1: An example from the Idiom dataset

Each entry in the dataset comprises three fields: (a) idiom — the surface form of the Bangla idiom, (b) meaning — a set of Bangla glosses capturing both primary and secondary interpretations, and (c) sentence — a Bangla sentence illustrating the idiom in natural usage. The final dataset con-tains 10,822 unique idiom forms, of which 2,624 exhibit multiple meanings. In addition, 4,772 entries are accompanied by contextual sentences. Table 1 presents a representative example.

[2] https://bn.wikipedia.org/wiki/বাংলাবাগধারারতালিকা
[3] https://www.english-bangla.com/lessons/bagdhara

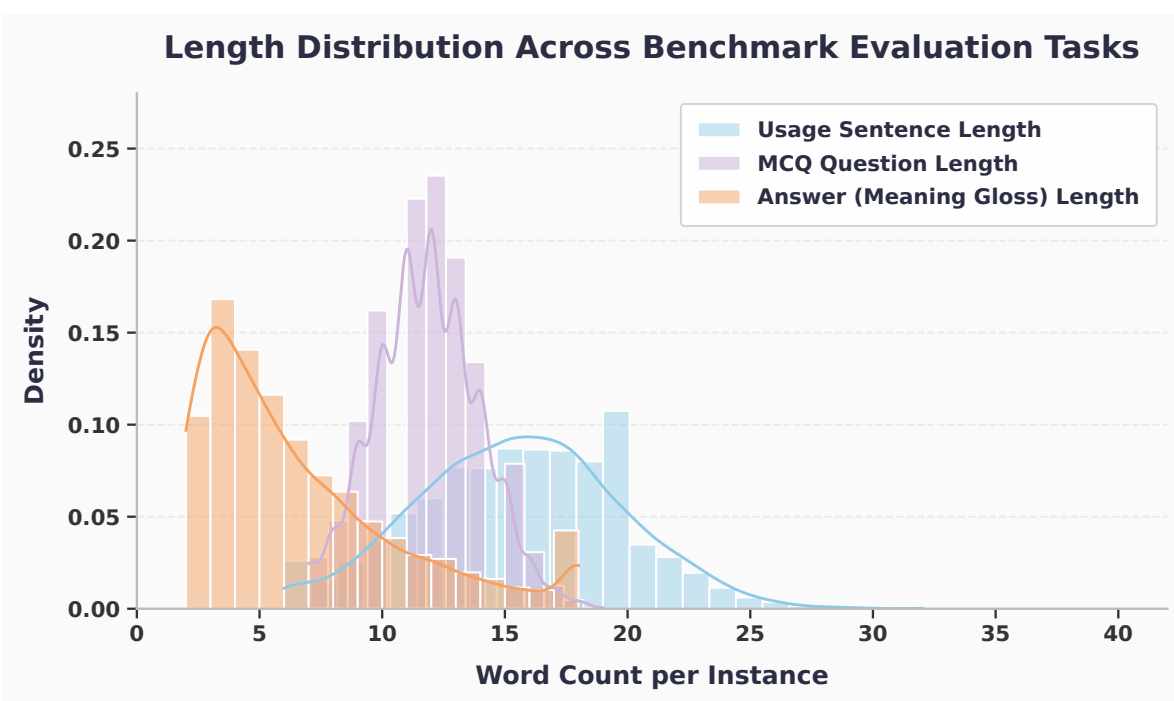

Figure 2: Word-count length distributions

Building on 10,822 idiomatic examples, we construct a multiple-choice (MCQ) dataset for Bangla idiom meaning detection, where each instance requires selecting the correct meaning of a given idiom from four candidate options. Since the dataset provides each idiom paired with its ground-truth meaning(s), we generate distractor options by randomly sampling meanings from other idioms and shuffling them together with the correct option(s) to form balanced MCQ sets. This randomization ensures variability while preventing positional bias across options. As several idioms admit multiple valid interpretations, we further formulate a multi-answer setting in which more than one option may be correct for a given item. The resulting dataset consists of 10,913 single-answer instances and 38,688 multi-answer instances. Table 3 summarizes the key statistics and characteristics of the proposed Bangla idiom dataset. As shown in Figure 2, word-count distributions separate cleanly across tasks: answer meanings remain concise (2–8 words), MCQ prompts cluster around formulaic templates (10–14 words), and usage sentences display the widest syntactic variation (8–35 words).

## 4 Experimental Setup

We define three tasks to evaluate the understanding of Bangla idioms by large language models (LLMs): **(1) paraphrasing, (2) idiom span detection, and (3) multiple-choice (MCQ) meaning detection**. The overarching goal is to assess the extent to which LLMs comprehend idiomatic expressions in Bangla—a low-resource language. We experimented with eight LLMs, including Kimi-K2-32B-Instruct (Bai et al., 2025), LLaMA-4-Scout-17B-Instruct (Meta, 2025), DeepSeek-R1-Distill-LLaMA-70B (Guo et al., 2025), GPT-OSS-20B, GPT-OSS-120B (OpenAI, 2025), Phi-4-Mini-Instruct (Abdin et al., 2024), Qwen3-32B (Team, 2024), and Gemini-2.5-Flash (Comanici et al., 2025), accessed through the Groq AI[4] API. The prompt templates used for each task are presented in Table 2. We evaluated the models on a subset of 1,000 samples from the dataset.

**Paraphrasing** For the paraphrasing task, we provided each model with a sentence containing an idiom and instructed it to generate an alternative version that preserves the original meaning. As ground truth, we constructed reference sentences by replacing the idiomatic span with its literal meaning. The model outputs were evaluated against these references using lexical overlap metrics (ROUGE) (Lin, 2004), semantic similarity metrics (BERTScore) (Zhang et al., 2020), and cosine similarity computed over embeddings from OpenAI, Gemini, and multilingual models such as LaBSE (Feng et al., 2022) and LASER (Artetxe and Schwenk, 2019).

**Idiom Span Detection** In the idiom span detection task, models were asked to identify the idiomatic portion within a given sentence. The dataset provides gold-standard idiom spans, which were used as ground truth. Performance was evaluated using n-gram overlap percentage and Levenshtein distance (Li and Liu, 2007) as similarity-based metrics.

**MCQ Meaning Detection** The MCQ meaning detection task involved two experimental setups: one where each question contained a single correct meaning, and another where multiple correct meanings were possible. This task was designed to assess whether LLMs can recognize all valid interpretations of idiomatic expressions. Evaluation was performed using accuracy as the metric.

## 5 Results and Analysis

### 5.1 Quantitative Analysis

Our first set of results pertains to the idiomatic dataset presented in Table 1, where large language models (LLMs) were evaluated under two prompting configurations: zero-shot and five-shot. We chose a 5-shot setup motivated by (Beauchemin et al., 2026), who showed that a 5-shot configuration serves as a highly stable proxy for evaluating an LLM's multiword reasoning boundaries. To prevent in-context contamination, the few-shot examples were drawn exclusively from a held-out pool that is entirely disjoint from the test set used in all evaluations. As expected, overall performance improved under the five-shot setting, reflecting the advantage of in-context examples for this task. As detailed in Table 4 for the paraphrasing evaluation, *Phi-4-mini-instruct* achieved the highest lexical overlap scores under the five-shot setup (ROUGE-1 of 0.63 and ROUGE-2 of 0.50), indicating its strong capacity to generate paraphrases that closely approximate the reference texts.

[4]https://groq.com/

| Task | Prompt |
|---|---|
| Paraphrasing | Paraphrase the following idiomatic Bengali sentence while keeping the original meaning unchanged: "চাকরী না পেয়ে আমার শাপে বর হয়েছে, ব্যবসা করে লক্ষ্মী পেয়েছি". Just output the paraphrased sentence and nothing more than that. No explanations needed at all. |
| Span-detection | Detect idiom from this Bengali sentence: "সাহস হারালে জীবনের এই অগ্নি পরীক্ষায় উত্তীর্ণ হবে কি করে?". Just output the extracted idiom and nothing more than that. No explanations needed at all. |
| mcq-single | Which one is the correct meaning of the Bengali idiom "'কাঁঠালের আমসত্ত্ব'? Find just one single option: "(1) মাঝে মাঝে (2) অদর্শনীয় বস্তু (3) অসম্ভব বস্তু (4) কোনো মধ্যস্থের সাহায্য না নিয়ে সোজাসুজি. Just output the correct option number and nothing more than that. |
| mcq-multiple | Which are the correct meanings of the Bengali idiom 'গোলেমালে চণ্ডীপাঠ'? Output all the correct options: (1) 'ঠিক না বেঠিক পড়া ধরা পড়ে না'-এই সুযোগে কাজে ফাঁকি (2) গণ্ডায় এণ্ডা মেলানো (3) মানুষের বিশ্বাস নষ্টচন্দ্র দর্শনে পাপ হয়। (4) অনেকে থাকার সুযোগে কাজে ফাঁকি দেওয়া. Just output the correct option numbers separated by comma and nothing more than that. |

Table 2: Prompt designs for various tasks. Correct answers are underlined for clarity but not provided.

| Statistic | Value |
|---|---|
| Total idioms | 10,822 |
| Entries with usage sentence | 4,772 (44.1%) |
| Entries with multiple meaning | 2,624 |
| Mean meanings per entry | 1.23 |
| Max meanings per entry | 16 |
| MCQ with single answers | 10,913 |
| MCQ with multiple answers | 38,688 |

Table 3: Dataset statistics

For semantic similarity–based metrics, performance varied across embedding configurations. This variation aligns with prior research (Mahajan et al., 2024) highlighting the limitations of surface sentence embeddings in fully capturing idiomatic equivalence. Notably, *GPT-OSS-20b* and *Phi-4-mini-instruct* delivered the strongest overall performance across metrics, while *Gemini-2.5-Flash* obtained the highest similarity score when OpenAI embeddings (text-emb-3-large) were applied (0.84). Considering both lexical overlap and embedding-based measures, *Phi-4-mini-instruct* achieves the highest overall average score (0.73), making it the most effective model for idiomatic sentence paraphrasing in Bangla.

Extending this analysis, we next examined LLM performance on the idiom span detection task under the same prompting conditions. Evaluation was based on two primary metrics: the percentage of unigram overlap within the identified idiomatic spans and the Levenshtein distance (LD) between predicted and reference spans. As summarized in Table 5, *Kimi-K2-32b-instruct* achieved the highest performance in the five-shot setting, yielding a unigram overlap of 48.25% and the lowest Levenshtein distance of 6.27. Consistent with the paraphrasing task, most models benefited from in-context demonstrations, further indicating that LLMs leverage example-based prompting effectively for extracting idiomatic spans.

Finally, we assessed model performance on idiom meaning identification using our synthetic MCQ dataset. In this task, models were prompted to select the correct meaning for each idiomatic expression, evaluated under both single-meaning and multi-meaning setup conditions. The detailed results are shown in Table 6. *Gemini-2.5-flash* achieved the highest accuracy across both subtasks, scoring 0.76 on single-answer detection and 0.55 on multi-answer detection. These results suggest its strong capability in capturing fine-grained semantic interpretations and disambiguating idiomatic senses in low-resource settings.

Overall, our analysis highlights notable variability in LLM performance across idiom-related tasks. While *Phi-4-mini-instruct* demonstrates strong proficiency in paraphrasing idiomatic sentences, *Kimi-K2-32b-instruct* excels in idiom span detection, and *Gemini-2.5-Flash* performs best in meaning identification for Bangla. These results suggest that no single model uniformly excels across all subtasks, underscoring the multifaceted nature of idiomatic understanding in Bangla.

| Model | Fewshot Setting | ROUGE-1 | ROUGE-2 | ROUGE-L | BERTScore | Cosine Similarity text-emb -3-large | gemini-emb -001 | LaBSE | LASER | AVG |
|---|---|---|---|---|---|---|---|---|---|---|
| Kimi-K2-32b-instruct | 0-shot | 0.38 | 0.09 | 0.38 | 0.76 | 0.57 | 0.81 | 0.73 | 0.80 | 0.57 |
| | 5-shot | 0.40 | 0.16 | 0.37 | 0.78 | 0.63 | 0.82 | 0.80 | 0.83 | 0.60 |
| LLaMA-4-17b-instruct | 0-shot | 0.41 | 0.22 | 0.37 | 0.85 | 0.60 | 0.81 | 0.76 | 0.77 | 0.60 |
| | 5-shot | 0.47 | 0.25 | 0.39 | 0.80 | 0.62 | 0.87 | 0.79 | 0.78 | 0.62 |
| DeepSeek-r1-70b | 0-shot | 0.47 | 0.17 | 0.44 | 0.74 | 0.70 | 0.81 | 0.70 | 0.71 | 0.59 |
| | 5-shot | 0.51 | 0.19 | 0.51 | 0.80 | 0.61 | 0.76 | 0.76 | 0.75 | 0.61 |
| GPT-OSS-20b | 0-shot | 0.60 | 0.30 | **0.56** | **0.88** | 0.77 | 0.85 | **0.88** | 0.85 | 0.71 |
| | 5-shot | 0.52 | 0.40 | 0.53 | **0.88** | 0.75 | **0.92** | **0.88** | 0.86 | 0.72 |
| GPT-OSS-120b | 0-shot | 0.54 | 0.18 | 0.43 | 0.83 | 0.72 | 0.88 | 0.84 | 0.81 | 0.65 |
| | 5-shot | 0.55 | 0.30 | 0.47 | 0.87 | 0.73 | 0.90 | 0.80 | 0.84 | 0.68 |
| Phi-4-mini-instruct | 0-shot | 0.22 | 0.02 | 0.26 | 0.74 | 0.68 | 0.79 | 0.68 | 0.73 | 0.52 |
| | 5-shot | **0.63** | **0.50** | **0.56** | 0.83 | 0.70 | 0.90 | 0.82 | **0.89** | **0.73** |
| Qwen3-32b | 0-shot | 0.55 | 0.24 | 0.46 | 0.74 | 0.75 | 0.80 | 0.69 | 0.68 | 0.61 |
| | 5-shot | 0.58 | 0.31 | 0.55 | 0.76 | 0.76 | 0.74 | 0.80 | 0.76 | 0.66 |
| Gemini-2.5-flash | 0-shot | 0.41 | 0.16 | 0.40 | 0.77 | 0.80 | 0.80 | 0.73 | 0.77 | 0.61 |
| | 5-shot | 0.57 | 0.26 | 0.54 | 0.80 | **0.84** | 0.84 | 0.84 | 0.87 | 0.70 |

Table 4: Paraphrasing task results across various LLMs under 0-shot and 5-shot settings. The highest score for each metric is highlighted in bold.

| **Model** | **Fewshot Setting** | **% Unigram Overlap** | **Levenshtein Distance (LD)** |
|---|---|---|---|
| Kimi-K2-32b-instruct | 0-shot | 37.66 | 8.21 |
| | 5-shot | **48.25** | **6.27** |
| LLaMA-4-17b-instruct | 0-shot | 32.12 | 13.34 |
| | 5-shot | 47.61 | 13.78 |
| DeepSeek-r1-70b | 0-shot | 26.54 | 10.47 |
| | 5-shot | 38.56 | 8.36 |
| GPT-OSS-20b | 0-shot | 34.23 | 11.36 |
| | 5-shot | 37.04 | 8.54 |
| GPT-OSS-120b | 0-shot | 36.07 | 9.62 |
| | 5-shot | 44.20 | 8.25 |
| Phi-4-mini-instruct | 0-shot | 11.00 | 9.30 |
| | 5-shot | 21.08 | 9.93 |
| Qwen3-32b | 0-shot | 39.39 | 7.55 |
| | 5-shot | 40.97 | 6.64 |
| Gemini-2.5-flash | 0-shot | 35.23 | 9.90 |
| | 5-shot | 42.68 | 7.91 |

Table 5: Results across various LLMs for the idiomatic span detection task under 0-shot and 5-shot settings. The best scores per metric are highlighted in bold.

**Finding:** Idiomatic understanding in Bangla is a multifaceted challenge. LLM performance on Bangla idiom tasks varies substantially across subtasks, with different models excelling in different areas, indicating that no single model consistently performs best across paraphrasing, span detection, and meaning identification.

### 5.2 Human Evaluation

To rigorously evaluate the contextual nuance and semantic preservation of Bangla idioms, we perform a manual human evaluation focused exclusively on the paraphrasing task. Automated evaluation metrics are often insufficient to capture the linguistic, cultural, and idiomatic subtleties inherent in Bangla. Accordingly, three highly qualified annotators (native speakers as well) assess a randomly sampled subset of 25 idiomatic expressions per model across eight models, under both zero-shot and five-shot settings (400 samples in total). The outputs are rated on a continuous scale from 0 to 2, where 0 indicates complete loss of idiomatic meaning or nonsensical output, 1 denotes a partially correct or overly literal paraphrase, and 2 corresponds to a fully faithful and natural-sounding

| Model | Single Detection | Multiple Detection |
|---|---|---|
| Kimi-K2-32b-instruct | 0.71 | 0.52 |
| LLaMA-4-17b-instruct | 0.74 | 0.48 |
| DeepSeek-r1-70b | 0.19 | 0.08 |
| GPT-OSS-20b | 0.60 | 0.23 |
| GPT-OSS-120b | 0.63 | 0.24 |
| Phi-4-mini-instruct | 0.29 | 0.26 |
| Qwen3-32b | 0.49 | 0.28 |
| Gemini-2.5-flash | **0.76** | **0.55** |

Table 6: Accuracy across various LLMs for the MCQ meaning detection task.

| **Model** | **A1 vs A2** | **A2 vs A3** | **A1 vs A3** | **Score** |
|---|---|---|---|---|
| Gemini-2.5-flash | 0.644 | 0.614 | 0.668 | 0.642 |
| DeepSeek-r1-70b | 0.608 | 0.635 | 0.636 | 0.626 |
| LLaMA-4-17b-instruct | 0.667 | 0.651 | 0.563 | 0.627 |
| Kimi-K2-instruct | 0.585 | 0.687 | 0.601 | 0.624 |
| GPT-OSS-20b | 0.589 | 0.783 | 0.501 | 0.624 |
| GPT-OSS-120b | 0.645 | 0.620 | 0.648 | 0.638 |
| Qwen3-32b | 0.631 | 0.754 | 0.824 | 0.737 |
| Phi-4-mini-instruct | 0.633 | 0.755 | 0.639 | 0.676 |
| **Average** | **0.625** | **0.687** | **0.635** | **0.65** |

Table 7: Inter-annotator Agreement (Cohen's Kappa) across models.

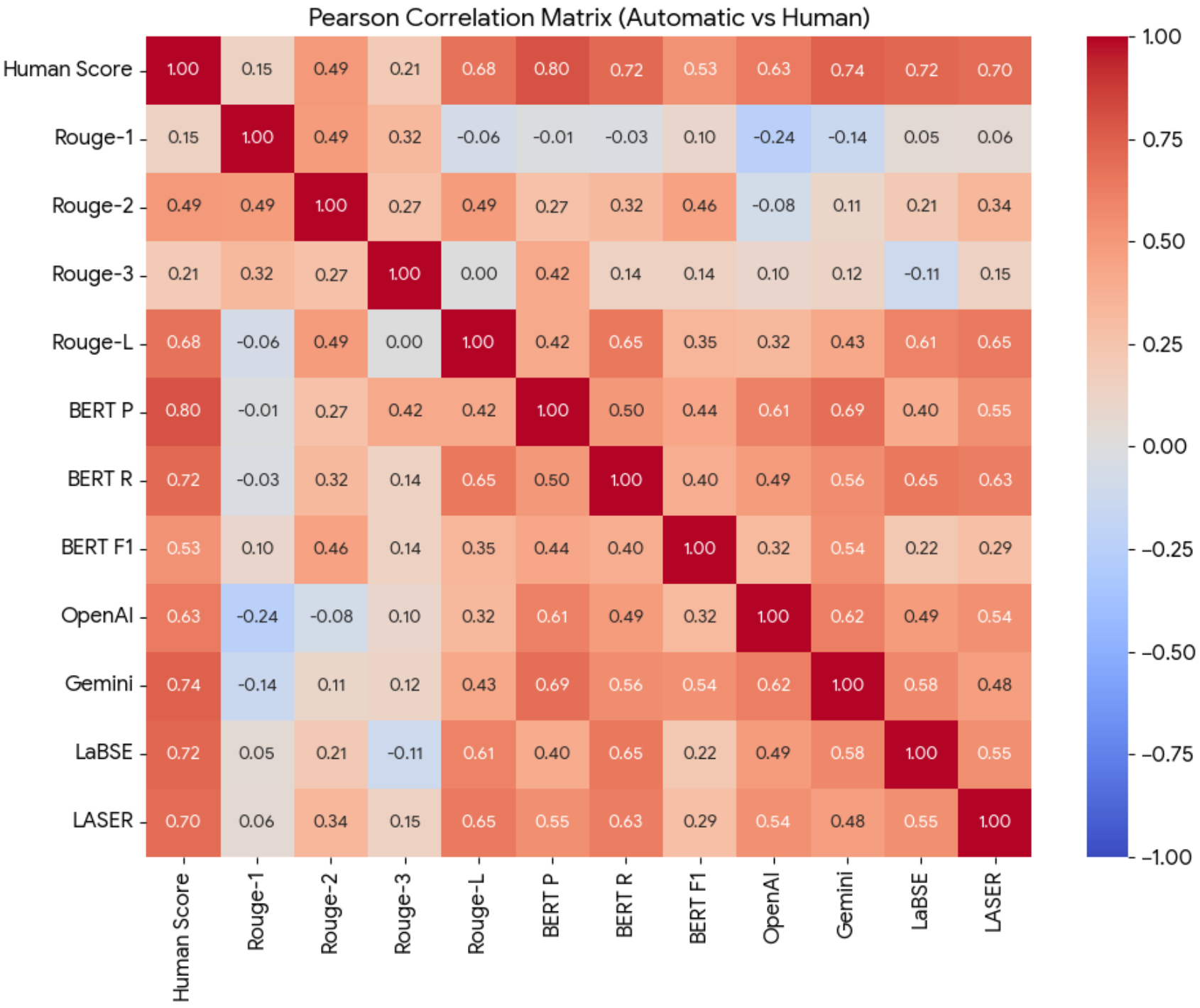

Figure 3: Pearson Correlation Matrix illustrating the linear relationship between automated evaluation metrics and human annotation scores.

paraphrase.

#### 5.2.1 Inter-annotator Agreement

To ensure the validity of our manual grading process, we calculated the inter-annotator agreement using pairwise weighted Cohen's Kappa scores across all three annotators (A1 vs. A2, A2 vs. A3, and A1 vs. A3). The annotations exhibited substantial agreement across the board. The total overall average Kappa score across all models and annotators stood at a robust 0.65.

**Highest Agreement.** The evaluators showed the highest consensus when grading the outputs of qwen3-32b, achieving an overall Kappa score of 0.737. This indicates that the outputs generated by this model were straightforward and unambiguous to grade.

**Lowest Agreement.** Conversely, kimi-k2-instruct and gpt-oss-20b tied for the lowest annotator agreement at 0.624, suggesting their paraphrased variations were slightly more ambiguous or prone to subjective interpretation by the native speakers.

#### 5.2.2 Human Assessment Scores

The raw human annotation scores reveal distinct differences in model capabilities regarding Bengali idiom comprehension. We averaged the human scores for each model out of the maximum possible score of 2.0.

| Model | 0-Shot | 5-Shot | Average |
|---|---|---|---|
| Gemini-2.5-flash | 1.417 | 1.333 | 1.375 |
| DeepSeek-r1-70b | 0.625 | 0.736 | 0.681 |
| LLaMA-4-17b-instruct | 0.917 | 0.917 | 0.917 |
| Kimi-K2-32b-instruct | 1.542 | 1.333 | **1.438** |
| GPT-OSS-20b | 0.875 | 0.889 | 0.882 |
| GPT-OSS-120b | 0.736 | 0.903 | 0.819 |
| Qwen3-32b | 0.458 | 0.708 | 0.583 |
| Phi-4-mini-instruct | 1.056 | 1.236 | 1.146 |

Table 8: Average human annotation scores across different models by different Prompt Settings.

**Top Performers.** The kimi-k2-instruct model achieved the highest overall average score of 1.438, closely followed by gemini-2.5-flash at 1.375. Interestingly, both of these leading models peaked during the zero-shot prompting phase (scoring 1.542 and 1.417, respectively) and slightly regressed when provided with five-shot examples (1.333 for both).

**Few-Shot Beneficiaries.** Conversely, several smaller or distilled models benefited heavily from the in-context learning provided by the five-shot prompts. phi-4-mini-instruct improved from 1.056 (zero-shot) to 1.236 (five-shot), and deepseek-r1-distill-llama-70b jumped from 0.625 to 0.736.

**Lowest Performers.** The qwen3-32b model struggled the most to capture the nuanced meaning of the idioms, returning the lowest overall average score of 0.583.

> **Finding:** Model performance varies across prompting settings, with leading models performing best in zero-shot, smaller models benefiting from few-shot examples, and Qwen3-32B showing the weakest overall performance.

#### 5.2.3 Validating Automatic Metrics against Human Judgement

To determine which automated metrics are the most reliable proxies for true human evaluation in Bengali paraphrase tasks, we computed the Pearson Correlation Coefficient ($r$) between the automated scores and the segment-level human annotations.

Our correlation analysis highlights a significant disparity between traditional surface-level metrics and modern embedding-based evaluators:

**Strongest Correlations.** Embedding-based metrics demonstrated the highest linear alignment with human judgment. **BERTScore Precision (BERT P)** yielded the strongest overall correlation ($r = 0.80$), followed by the **Gemini Embedding** similarity ($r = 0.74$) and **BERTScore Recall (BERT R)** ($r = 0.72$). **LaBSE** also proved to be a highly reliable multilingual metric, achieving a Pearson correlation of $r = 0.72$.

**Weakest Correlations.** Traditional $n$-gram overlap metrics proved highly ineffective at capturing the semantic preservation of idioms. **ROUGE-1** ($r = 0.15$) and **ROUGE-3** ($r = 0.21$) showed virtually no meaningful correlation with human preference, emphasizing that surface-level lexical matching fails to reward models that accurately paraphrase the *meaning* of an idiom using entirely different vocabulary.

> **Finding:** The evaluation of Bengali idiomatic paraphrasing should prioritize embedding-based semantic metrics (e.g., BERTScore, LaBSE) over lexical metrics (e.g., ROUGE), as they better align with human judgments.

## 6 Error Analysis and Limitations of Open-Source LLMs for Bangla Idiom Understanding

### 6.1 Failures in Bangla Pre-trained LLMs

We initially evaluated several Bangla pre-trained large language models (LLMs), including *TituLLM-3B* (Nahin et al., 2025), *BanglaLLaMA3-8B* (Zehady, 2025), and *OdiaGenAI-Bengali-LLaMA7B* (Parida et al., 2023). However, these models consistently failed to produce coherent or contextually relevant responses even for simple conversational prompts. For example, when prompted with “আমি কি দয়া করে জানতে পারি আপনি কে বলছেন?” (*May I please know who are you?*), one model generated unrelated and repetitive text such as “আমি একটা ছোট্ট দোকানে ঢুকলাম। সেখানে একটা সুন্দর সাদা রঙের টেবিল ছিল।...” (*I entered a small shop. There was a beautiful white table there...*). Notably, this failure occurred outside the idiom understanding task itself and instead reflected a broader inability to maintain semantic coherence in general Bangla text generation. Due to such persistent irrelevance and instability in outputs, we excluded these Bangla-specific pre-trained models from the final benchmarking suite.

> **Finding:** Existing Bangla pre-trained LLMs struggle to maintain semantic coherence even in simple conversational settings, limiting their applicability to downstream idiom understanding tasks.

### 6.2 Idiom Span Detection: Zero-Shot vs. Few-Shot Prompting

Few-shot prompting is commonly expected to improve model performance by providing structured demonstrations. However, in our Bangla idiom span detection experiments, we observed that additional examples occasionally introduced hallucinations and degraded performance. This behavior was particularly evident in open-source models such as Qwen3-32B.

For instance, given the sentence “‘ভ্রূবিলাস শেখে নাই কারা সেই নারী’ - রবীন্দ্রনাথ ঠাকুর,” (*Who is the woman who has not learned the play of eyebrows? – Rabindranath Tagore*), the zero-shot setup extracted “ভ্রূবিলাস শেখে নাই” (*has not learned the play of eyebrows*), which overextended the span but still included the correct idiom “ভ্রূবিলাস” (play of eyebrows). In contrast, the 5-shot prompting setup identified only "শেখে নাই" (*has not learned*), omitting the core idiomatic expression entirely. Although the zero-shot prediction remained imperfect, it was substantially closer to the ground truth than the few-shot variant. This suggests that, in certain settings, few-shot demonstrations may inadvertently confuse models toward structurally similar yet semantically irrelevant phrases.

**Finding:** Few-shot prompting can sometimes degrade Bangla idiom span detection performance by introducing hallucinated or semantically irrelevant spans.

### 6.3 Over-Extraction in Idiom Span Detection

A recurring limitation across open-source LLMs in the span-based Bangla idiom detection task is the tendency to extract overly long spans that include non-idiomatic surrounding words. This over-extraction issue becomes especially prominent when the idiom appears within a broader noun phrase or figurative construction.

For example, in the sentence "রাজনীতির ময়দানে শুধু ঘোড়েলদের ঘোরাফেরা" (*In the field of politics, only the clever ones roam around*), the correct idiomatic expression is "ঘোড়েল" (*clever*). However, several open-source models, including Qwen3-32B, Meta LLaMA-4-17B, DeepSeek-R1-Distill-LLaMA-70B, and GPT-OSS-20B, incorrectly extracted the extended span "ঘোড়েলদের ঘোরাফেরা" (*the roaming of the clever ones*).

In contrast, the closed-source model Gemini-2.5-Flash successfully identified the precise idiomatic boundaries in such cases, demonstrating stronger fine-grained semantic understanding. These findings indicate that although open-source LLMs show promising capabilities, they still lag behind proprietary models in accurately capturing idiomatic spans and figurative language phenomena in low-resource languages such as Bangla.

**Finding:** Open-source LLMs frequently over-extract Bangla idiomatic spans, whereas proprietary models demonstrate stronger boundary-level semantic understanding.

### 6.4 Difficulty in Multi-Label MCQ Reasoning

The multiple-choice question (MCQ) task in our benchmark requires models to identify all valid meanings of a Bangla idiom from four candidate options. Unlike standard single-answer MCQs, this task involves multi-label semantic reasoning, where multiple options may simultaneously be correct, and models are expected to output only the indices of the correct choices in a comma-separated format.

For example, given the idiom "অকর্মার ধাড়ি" (*a useless person*), the candidate meanings are: (1) বিবাহকালে শুভদৃষ্টি (*The auspicious exchange of glances during the wedding)*, (2) ধানকাঠের মই ( *a useless person*), (3) খোদার খাসি (*A carefree person*), and (4) অগাকান্ত (*ignorant person*). The correct answers are options 2, 3, and 4. However, none of the evaluated open-source models successfully identified the complete set of correct answers, with option (2) ধানকাঠের মই, being most frequently omitted where the literal translation is *paddy straw ladder*. By contrast, Gemini-2.5-Flash correctly selected all valid options while also adhering to the required output format. This highlights a substantial gap between open-source and proprietary models in handling nuanced multi-label semantic reasoning tasks in Bangla, particularly when meanings involve culturally grounded interpretations.

**Finding:** Open-source LLMs struggle with multi-label semantic reasoning in Bangla idiom MCQs, especially when multiple culturally grounded meanings must be identified simultaneously.

## 7 Conclusion

In this paper, we introduced the first large-scale benchmark dataset of Bangla idioms, along with an additional synthetic multiple-choice question (MCQ) dataset designed for idiom meaning detection. We conducted a series of experiments using recent LLMs across multiple idiom-related tasks, including paraphrasing, idiom span detection, and meaning identification, employing various prompting strategies. Our findings reveal notable variability in LLM performance, indicating that no single model performs optimally across all tasks. We believe our dataset and analyses provide a foundation for advancing idiomatic understanding in LLMs, particularly for low-resource languages like Bangla. Further linguistic analyses enabled by the dataset, such as translation challenges and vocabulary complexity, are beyond the scope of this work and left for future research.

## Limitations

While the dataset is broad and carefully curated, some idioms may have been missed unintentionally, and future work can further expand the dataset. Due to computational constraints, experiments were conducted on a subset of 1,000 samples per setting, which may reduce evaluation granularity but still captures overall performance trends. Additionally, only 4,772 out of 10,822 idiomatic expressions (44.1%) are accompanied by contextual usage sentences, limiting their applicability for context-dependent tasks. However, these sentences were directly collected from original sources and represent human-written examples, while missing contexts were unavailable in the source materials. The synthetic MCQ benchmark also relies on randomly sampled meanings as distractors, which may allow models to solve questions through elimination; future work can explore more challenging distractor construction strategies. Finally, budget constraints limited the inclusion of some commercially available or high-cost LLMs in our evaluation. These limitations do not diminish our central contribution: a large-scale Bangla idiom dataset and a comprehensive evaluation of LLM capabilities on idiom-related tasks, including paraphrasing, span identification and meaning detection.

# A Appendix

## A.1 Data Licensing and Source Redistribution Permissions

All text sources used to construct the Bangla idiom benchmark are distributed under open, public, or Creative Commons frameworks, permitting their use, adaptation, and redistribution for educational and non-commercial research purposes. The extracted source annotations, vocabulary phrases, and contextual sentences fall under permissive reuse guidelines for academic evaluation tasks. A comprehensive breakdown of the licensing framework for each source category is summarized in Table 9. Specifically, our data collection pipeline draws from four primary categories:

| Data Source Category | Primary Content Origin | Applicable Distribution License / Term | Permissible Academic Use |
|---|---|---|---|
| Online Encyclopedias | Bangla Wikipedia | Creative Commons Attribution-ShareAlike (CC BY-SA 4.0) | Copying, remixing, and redistribution permitted with attribution under the same license. |
| Online Lexicons | Public Reference Dictionaries | Open Educational Resources (OER) Terms / Public Domain | Permitted for vocabulary mapping, non-commercial research, and text parsing evaluation. |
| National Curriculum | NCTB Grammar Textbooks | Public Distribution (Free Government Textbooks) | Openly accessible and allowed for non-commercial educational analysis and replication. |
| Print Media | Bangla Newspapers | Fair Use / Fair Dealing (Section 52 of Bangladesh Copyright Act) | Permitted for non-commercial linguistic evaluation, syntactic parsing, and text benchmarking. |

Table 9: Source licensing and distribution permission mapping for the benchmark dataset.

**Online Encyclopedias:** Entries sourced from Bangla Wikipedia are licensed under the Creative Commons Attribution-ShareAlike 4.0 International (CC BY-SA 4.0) license, which permits copying, remixing, and redistribution under the same terms with proper attribution.

**Online Lexicons:** Vocabulary mappings and reference definitions sourced from public reference dictionaries fall under Open Educational Resources (OER) terms and public domain guidelines, allowing non-commercial research and computational parsing.

**National Curriculum:** Idiomatic examples and grammar rules derived from National Curriculum and Textbook Board (NCTB) grammar textbooks are freely distributed government educational materials, permitted for open academic analysis and reproducibility.

**Print Media:** Natural usage sentences extracted from Bangla newspapers are utilized under Fair Use and Fair Dealing provisions (e.g., Section 52 of the Bangladesh Copyright Act), which explicitly allow the reproduction of published works for non-commercial educational instruction, linguistic evaluation, and criticism.

### A.2 Data Collection Strategy

Our data collection strategy followed a two-stage digitization and curation framework:

**Digital/Online Sourcing:** For natively digital platforms like Bangla Wikipedia and established online lexicons, data was manually collated to build the core electronic lexicon.

**Print/Offline Sourcing:** For physical text resources, specifically the National Curriculum and Textbook Board (NCTB) grammar books (বাংলা ভাষার ব্যাকরণ) and contemporary print newspapers, we captured high-resolution source images. These images were processed through an Optical Character Recognition (OCR) pipeline optimized for the Bengali script to convert print text into editable strings.

Following extraction, the authors (Bangla native speakers) thoroughly audited all entries to fix spelling inconsistencies and filter out duplicates. Human annotators also identified idioms with multiple senses, which are stored as a single list under the corresponding idiom entry. To maximize usability for future benchmarking, all verified artifacts were structured into a standardized JSON schema.